\documentclass[letterpaper, 10 pt, journal, twoside]{IEEEtran}

\usepackage{enumitem}
\usepackage{amsmath,amsfonts}
\usepackage{amsthm}

\newtheorem{theorem}{Theorem}

\usepackage{multirow}
\usepackage{graphics}
\usepackage{graphicx}
\usepackage{algorithm}
\usepackage{algcompatible}
\usepackage{subfigure}
\usepackage{bbm}
\usepackage{hyperref}
\hypersetup{
    colorlinks=true,
    linkcolor=cyan,
    filecolor=magenta,      
    urlcolor=cyan,
}
\usepackage{amssymb}

\def\BibTeX{{\rm B\kern-.05em{\sc i\kern-.025em b}\kern-.08em
    T\kern-.1667em\lower.7ex\hbox{E}\kern-.125emX}}

\begin{document}
\title{Interactive Semi-Parametric On-Track Learning for Vehicle Dynamics in Autonomous Racing}

\author{Jungbin Kim, Youngim Nam, and Cheolhyeon Kwon%
\thanks{This work has been submitted to the IEEE for possible publication.
Copyright may be transferred without notice, after which this version
may no longer be accessible.}%
\thanks{Jungbin Kim and Youngim Nam contributed equally to this work. (Corresponding author: Cheolhyeon Kwon.)}%
\thanks{The authors are with the Department of Mechanical Engineering, Ulsan National Institute of Science and Technology, Ulsan 44919, Republic of Korea (e-mail: rlawjdqls1212@unist.ac.kr; nyi0944@unist.ac.kr; kwonc@unist.ac.kr).}}

\maketitle
\begin{abstract}
Autonomous racing requires an accurate vehicle dynamics model near the limits of handling, yet a mismatch between the nominal and actual dynamics inevitably remains.
To reduce this mismatch, existing methods learn from on-track data, i.e., data collected while the vehicle drives on the track under racing conditions, by adapting the physical parameters of the nominal model or by learning a residual that captures the remaining mismatch.
However, these on-track learning methods typically handle either the physical parameters or the residual separately, which restricts their ability to resolve the mismatch.
To address this limitation, we propose an interactive semi-parametric learning framework that learns both the physical parameters and the residual under a single shared prediction-error objective.
Since updating one changes the learning target of the other, they are recursively updated in turn, each guided by the other's latest update.
Each update is then accepted only if it does not increase the shared objective, so that the objective is guaranteed to be non-increasing and convergent over the recursion.
Numerical simulations and hardware experiments show that the proposed framework outperforms the baselines.
Compared with the nominal model, it reduces the one-step prediction errors of lateral velocity and yaw rate by 90.0\% and 81.6\% in simulation and by 54.0\% and 59.4\% on hardware.
\end{abstract}

\begin{IEEEkeywords}
Autonomous Racing, On-Track Learning, Semi-Parametric Learning, System Identification
\end{IEEEkeywords}

\section{Introduction}
\begin{figure}[!t]
    \centering
\includegraphics[width=0.45\textwidth]{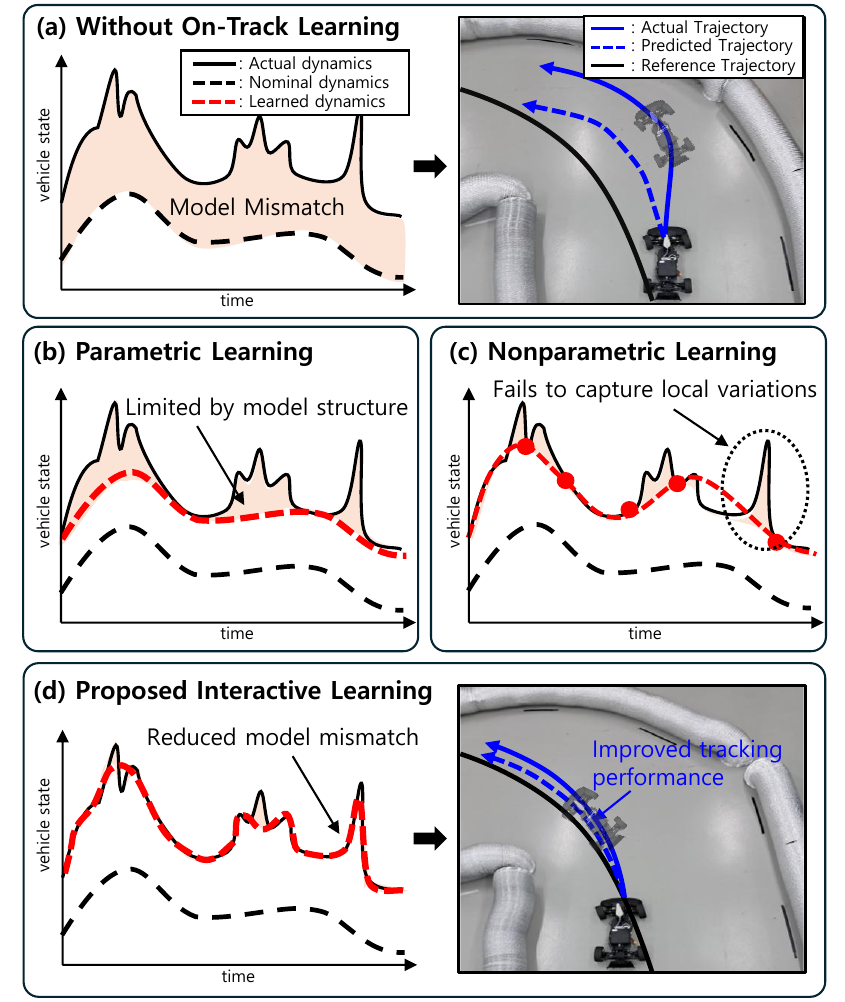} 
    \caption{Illustration of the learning approaches: (a) Without learning from on-track data, the nominal dynamics deviate from the actual dynamics; (b) Parametric learning is limited by the model structure; (c) Nonparametric learning with a sparse Gaussian process fails to capture localized variations; (d) The proposed interactive learning updates both the physical parameters and the residual while accounting for their mutual dependence.}
    \label{fig.concept}
\end{figure}

\IEEEPARstart{I}{n} autonomous racing, a vehicle must complete a track as quickly as possible while operating near the limits of handling \cite{survey_racing}.
Doing so requires a controller built on an accurate dynamics model, yet the actual vehicle dynamics are hard to identify.
For instance, physical parameters, such as tire friction, vary with operating conditions, and complex nonlinear effects, such as lateral load transfer, are difficult to capture accurately \cite{survey_model}.
Traditionally, such parameters are obtained through steady-state calibration performed before racing \cite{offtrack1, offtrack2}, but the resulting values may not remain valid during racing.
With off-track calibration alone, these factors inevitably cause a mismatch between the nominal and actual dynamics (see Fig.~\ref{fig.concept}(a)), motivating efforts to learn vehicle dynamics from on-track data, i.e., data collected while the vehicle drives on the track under racing conditions.

One common approach is parametric learning, which identifies the physical parameters of a dynamics model \cite{param_adap_NLS, param_adap_NLS2}.
While this approach can mitigate the mismatch caused by inaccurate parameter values, its effectiveness is inherently constrained by the underlying model structure (see Fig.~\ref{fig.concept}(b)).
For computational efficiency, most studies rely on simplified models that fail to capture complex nonlinear effects, regardless of parameter accuracy \cite{param_adap_NN, param_adap_lla}.
Increasing model fidelity can address this limitation, but the additional parameters make reliable identification difficult with limited on-track data \cite{param_adap_local_min}.

Instead, nonparametric learning retains a simplified parametric model and captures the unmodeled dynamics as a residual, yielding a semi-parametric dynamics model \cite{residual_gp, residual_gp2}.
This residual is commonly learned using Gaussian process (GP) regression due to its flexibility in modeling nonlinear functions. 
However, standard GP regression scales poorly with dataset size, making its use challenging in real-time control.
Sparse approximations address this issue using a small set of inducing points \cite{sparsegp}, but they often fail to capture localized variations in the residual \cite{sparsegp_mpc} (see Fig.~\ref{fig.concept}(c)).
This difficulty is further compounded because the underlying parametric model is not updated, leaving the sparse GP to compensate for the entire mismatch.
In essence, existing dynamics learning methods in racing typically pursue either parametric learning of the physical parameters or nonparametric learning of the residual in isolation, leaving part of the mismatch unresolved.

A straightforward approach to combining parametric and nonparametric learning is to apply them sequentially, i.e., to adapt the parametric model first and then train the residual model on what remains unmatched.
However, the two models are mutually dependent, so this sequential learning can lead to biased estimates on both sides.
In this case, the parameters are distorted to compensate for the unmodeled dynamics, and the residual model then learns from this distorted parametric model~\cite{joint_joint}.
Re-adapting the parameters using the learned residual, in turn, changes the residual target and can leave the residual model outdated~\cite{joint_simultaneous}.
For robot manipulators, a few studies have addressed this dependence by analytically adjusting the residual model according to parameter updates~\cite{joint_simultaneous} or by estimating both models in a single regression~\cite{joint_joint}.
Both approaches, however, exploit the linearity of rigid-body dynamics in the inertial parameters, limiting their direct extension to nonlinearly parameterized vehicle dynamics.

To address these challenges, this paper proposes an interactive framework for semi-parametric learning from on-track data in autonomous racing.
The framework learns both the physical parameters and the residual model under a single shared objective, namely the prediction error of the resulting semi-parametric model.
Since the nonlinear parameterization of the dynamics and the nonparametric nature of the residual model make this objective difficult to minimize directly, the two models are recursively updated in turn.
In doing so, each model guides the other with its latest update, accounting for their mutual dependence.
As a result, neither model has to compensate for the part of the mismatch that the other can explain, unlike when each is learned in isolation.

Specifically, once the data are collected, the physical parameters are adapted via nonlinear least-squares (NLS) optimization~\cite{param_adap_NN} given the current residual model.
The residual model is then learned via sparse GP regression~\cite{svgp} on residual targets recomputed from the adapted parameters.
Each parameter update in this recursion shifts the residual targets by an amount that depends on the parameter change.
The sparse GP must therefore be updated by a matching amount, without increasing the shared prediction-error objective.
To this end, we introduce an adaptive learning rate and prediction-error-based model selection.
Since neither model update is allowed to increase the shared objective, the objective is provably non-increasing and convergent over the recursion.
Given the same limited on-track data, the resulting model achieves higher prediction accuracy than either model learned alone.
The overall contributions are summarized as follows:
\begin{itemize}[nosep]
   \item A semi-parametric dynamics learning framework that learns both the physical parameters and the residual model from on-track data in autonomous racing;

   \item An interactive recursive learning scheme in which the two models guide each other's updates under a shared objective, accounting for their mutual dependence;

   \item A sparse GP learning strategy in which an adaptive learning rate scales each GP update with the parameter change, and prediction-error-based model selection ensures that the objective is non-increasing and convergent over the recursion; and
   
   \item Evaluation through numerical simulations and real-world autonomous racing experiments, demonstrating improved dynamics prediction accuracy and lap time.
\end{itemize}

\section{Related Work}
\subsection{Parametric/Nonparametric Learning in Racing}
To obtain an accurate dynamics model, its physical parameters have traditionally been identified through steady-state calibration, which typically relies on a wide range of off-track data from circular (skidpad) driving~\cite{offtrack1, offtrack2}.
Collecting such data requires large open areas, so the calibration is performed before racing.
The identified parameters therefore may not reflect the tire and track conditions encountered during racing.
To address this limitation, on-track learning has been explored along two main directions: parametric learning of the physical parameters~\cite{param_adap_NLS} and nonparametric learning of the residual~\cite{residual_gp}.
The first direction focuses on identifying the physical parameters by minimizing prediction errors through NLS optimization.
Alternatively, \cite{param_adap_NN2} replaces the NLS problem with a neural network (NN) that identifies the tire coefficients and the yaw inertia within physically meaningful ranges.
Parametric learning, however, typically requires sufficiently rich on-track data, which are difficult to collect during racing.
To alleviate this difficulty, \cite{param_adap_NN} utilizes an NN-based residual model, but only to generate virtual steady-state data for NLS-based parametric learning.
Even with accurately identified parameters, the resulting model remains constrained by its structure and cannot capture unmodeled effects such as lateral load transfer.

The second direction adds a residual model to capture the mismatch that the parametric model cannot explain.
The residual model is typically learned from on-track data using a GP~\cite{residual_gp, residual_gp2} or an NN~\cite{residual_NN}, while the physical parameters are held at values identified before racing or specified in advance.
GPs are widely used because they provide uncertainty estimates along with the prediction, but their computational cost grows rapidly with the amount of data. 
\cite{residual_onlinegp} therefore adopts a sparse GP to run the controller in real time. 
However, with the parametric model fixed, the limited capacity of the sparse GP must also compensate for the error caused by inaccurate parameters.
Methods that update the residual model iteratively with newly collected data~\cite{residual_iterative, residual_iterative2} still keep the physical parameters unchanged, so this burden persists and can limit further gains in prediction accuracy.

\subsection{Semi-Parametric Learning in Robotics}
To improve prediction accuracy, some robotics studies have proposed learning both the physical parameters and the residual model, particularly for manipulators~\cite{joint_simultaneous, joint_up1}.
One approach proceeds sequentially, adapting the parameters first and then learning the residual model on what remains unmatched~\cite{joint_seq}.
Some methods update both models incrementally while identifying the physical parameters independently of the residual model~\cite{joint_up1}.
However, subsequent parameter updates change the residual target, potentially leaving the previously learned residual model inconsistent with the updated parametric model.
A straightforward way to restore consistency is to retrain the residual model from scratch, but this incurs additional computational cost.
To address this issue, \cite{joint_simultaneous} analytically adjusts the residual model according to the parameter change instead of retraining it, so that both models can be updated simultaneously.
Alternatively, \cite{joint_joint} estimates both models in a single regression, so that neither is learned against an outdated estimate of the other.
Nonetheless, both approaches exploit the fact that the rigid-body model of manipulators is linear in its inertial parameters.
These approaches therefore do not directly extend to autonomous racing, where the vehicle operates near the limits of handling and its dynamics depend nonlinearly on the physical parameters, especially through the tire model.
Overall, jointly learning the physical parameters and residual model from on-track data remains largely unexplored for the nonlinearly parameterized vehicle dynamics of autonomous racing.

\section{Problem Formulation}
\subsection{Parametric Dynamics Model}
\begin{figure}[!t]
    \centering
    \includegraphics[width=0.48\textwidth]{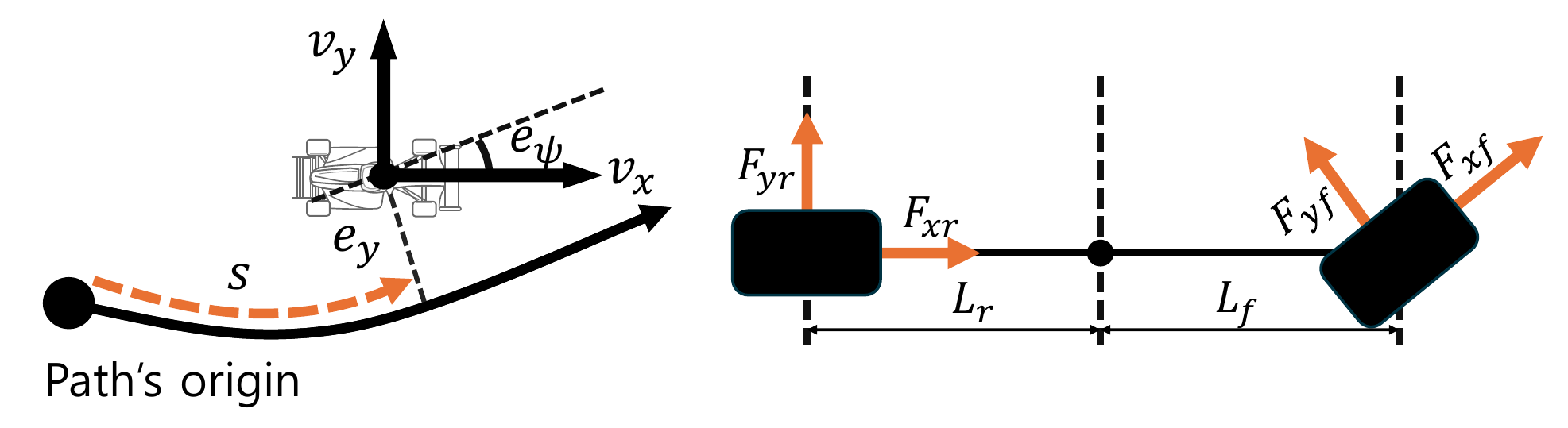}
    \caption{Schematic of the single-track model in the Frenet frame.}
    \label{fig:dyn}
\end{figure}

We base the parametric model on a nonlinear single-track model~\cite{Nominal} in the Frenet frame with respect to the track centerline (Fig.~\ref{fig:dyn}).
The state and input are defined as
\begin{equation*}
\mathbf{x}:=[\,s,\,e_y,\,e_\psi,\,v_x,\,v_y,\,\omega\,]^\top,\qquad
\mathbf{u}:=[\,a,\,\delta\,]^\top,
\end{equation*}
where $s$ is the progress along the centerline, $e_y$ and $e_\psi$ are the lateral and heading errors, $v_x$ and $v_y$ are the body-frame longitudinal and lateral velocities, and $\omega$ is the yaw rate. 
With the longitudinal acceleration $a$ and the steering angle $\delta$ as inputs, the continuous-time dynamics are
\begin{equation}\label{eq.continuous_dyn}
\begin{aligned}
&\dot{s} =\frac{v_x\cos e_\psi-v_y\sin e_\psi}{1-\kappa(s) e_y},\\
&\dot{e}_y=v_x\sin e_\psi+v_y\cos e_\psi,\\
&\dot{e}_\psi=\omega-\kappa(s)\,\frac{v_x\cos e_\psi-v_y\sin e_\psi}{1-\kappa(s) e_y},\\
&\dot{v}_x=a+\omega v_y-\frac{1}{m}\sum_{i\in\{f,r\}} F_{yi}\sin\delta_i,\\
&\dot{v}_y=-\omega v_x+\frac{1}{m}\sum_{i\in\{f,r\}} F_{yi}\cos\delta_i,\\
&\dot{\omega}=\frac{1}{I_z}\sum_{i\in\{f,r\}} \ell_i\,F_{yi}\cos\delta_i,
\end{aligned}
\end{equation}
where $m$ is the mass, $I_z$ is the yaw inertia, and $\kappa(s)$ is the track curvature.
The subscript $i\in\{f,r\}$ denotes the front and rear axles, with steering angles $\delta_f=\delta$ and $\delta_r=0$.
The signed distance from the center of gravity to axle $i$ is denoted by $\ell_i$, with $\ell_f=L_f>0$ and $\ell_r=-L_r<0$.
The lateral tire force at each axle follows a simplified Pacejka law~\cite{pacejka}:
\begin{equation*}
\begin{aligned}
&F_{yi}=\mu F_{zi}\sin\!\big(C\arctan(B\alpha_i)\big), \\
&\alpha_i=\delta_i-\arctan\!\Big(\tfrac{v_y+\ell_i \omega}{v_x}\Big),
\end{aligned}
\end{equation*}
where $\alpha_i$ is the slip angle, $F_{zi}$ is the normal load on axle $i$, and $B$, $C$, and $\mu$ are the stiffness, shape, and friction parameters.

Among these quantities, $m$ and $\ell_i$ can be obtained from static axle loads and vehicle geometry, whereas $I_z$ requires a dedicated test rig, and the tire parameters $B$, $C$, and $\mu$ vary with tire wear and track surface conditions.
We therefore define the unknown physical parameters as $\theta := [B, C, \mu, I_z]^\top$, which are adapted from on-track data within the set of physically plausible values $\Theta := [\theta_{\min}, \theta_{\max}]$.
Discretizing~\eqref{eq.continuous_dyn} yields the parametric model in terms of $\theta$:
\begin{equation}\label{eq.parametric}
\hat{\mathbf{x}}_{k+1} = f(\mathbf{x}_k, \mathbf{u}_k; \theta), \qquad \theta \in \Theta,
\end{equation}
where $\hat{\mathbf{x}}_{k+1}$ denotes the predicted state at the discrete time step $k+1$.

\subsection{Nonparametric Residual Model}
Even if $\theta$ were known exactly, the parametric model~\eqref{eq.parametric} could not fully capture the vehicle dynamics near the limits of handling.
For instance, the single-track model aggregates the individual wheel loads and lateral tire forces into a per-axle description, and thus does not explicitly account for lateral load transfer between the left and right wheels.
To compensate for the resulting mismatch, we introduce
a nonparametric residual model $g$ with a GP prior:
\begin{equation}\label{eq.residual_model}
g \sim \mathcal{GP}\big(0, K(\mathbf{z},\mathbf{z}')\big),
\end{equation}
where $\mathbf{z}$ is the regression input and $K$ denotes a matrix-valued covariance kernel.
Since the nominal modeling error is most significant in the lateral and yaw dynamics, the residual is restricted to $v_y$ and $\omega$~\cite{param_adap_NN, residual_iterative}:
\begin{equation*}
g(\mathbf{z}) = [\,0,\,0,\,0,\,0,\,g_{v_y}(\mathbf{z}),\,g_{\omega}(\mathbf{z})\,]^\top,
\end{equation*}
which corresponds to $K=\mathrm{diag}(0,0,0,0,k_{v_y},k_\omega)$ in~\eqref{eq.residual_model}, with scalar covariance kernels $k_{v_y}$ and $k_\omega$.
The regression input at time step $k$ is defined as
\begin{equation*}
\mathbf{z}_k := [\,v_{y,k},\,\omega_k,\,\delta_k,\,a_k\,]^\top.
\end{equation*}
The two nonzero components then approximate the part of the corresponding next states left unexplained by the parametric model:
\begin{equation}\label{eq.residual_target}
\begin{bmatrix} g_{v_y}(\mathbf{z}_k) \\ g_{\omega}(\mathbf{z}_k) \end{bmatrix}
\approx
\begin{bmatrix} v_{y,k+1} - f_{v_y}(\mathbf{x}_k,\mathbf{u}_k;\theta) \\ \omega_{k+1} - f_{\omega}(\mathbf{x}_k,\mathbf{u}_k;\theta) \end{bmatrix},
\end{equation}
where $v_{y,k+1}$ and $\omega_{k+1}$ denote the actual next states, and $f_{v_y}$ and $f_\omega$ denote the corresponding components of $f$.
Incorporating $g$ into the parametric model~\eqref{eq.parametric} yields the semi-parametric model:
\begin{equation}\label{eq.semiparametric}
\hat{\mathbf{x}}_{k+1} = f(\mathbf{x}_k,\mathbf{u}_k;\theta) + g(\mathbf{z}_k), \qquad \theta\in\Theta,\; g\in\mathcal{G},
\end{equation}
where $\mathcal{G}$ denotes the class of such residual models.

\subsection{Semi-Parametric Learning Problem}
The prediction accuracy of \eqref{eq.semiparametric} depends on both the physical parameters $\theta$ and the residual model $g$.
The two are inherently coupled: a change in $\theta$ alters the residual that $g$ must approximate, while a change in $g$ alters the mismatch that $\theta$ must explain.
To address this coupling, we formulate a semi-parametric learning problem that minimizes the prediction-error objective $J_{\mathcal{D}}$ over both $\theta$ and $g$:
\begin{equation}\label{eq.interactive_problem}
\min_{\theta\in\Theta,\;g\in\mathcal{G}}
J_{\mathcal{D}}(\theta,g).
\end{equation}

\begin{figure}[!t]
    \centering
\includegraphics[width=\columnwidth]{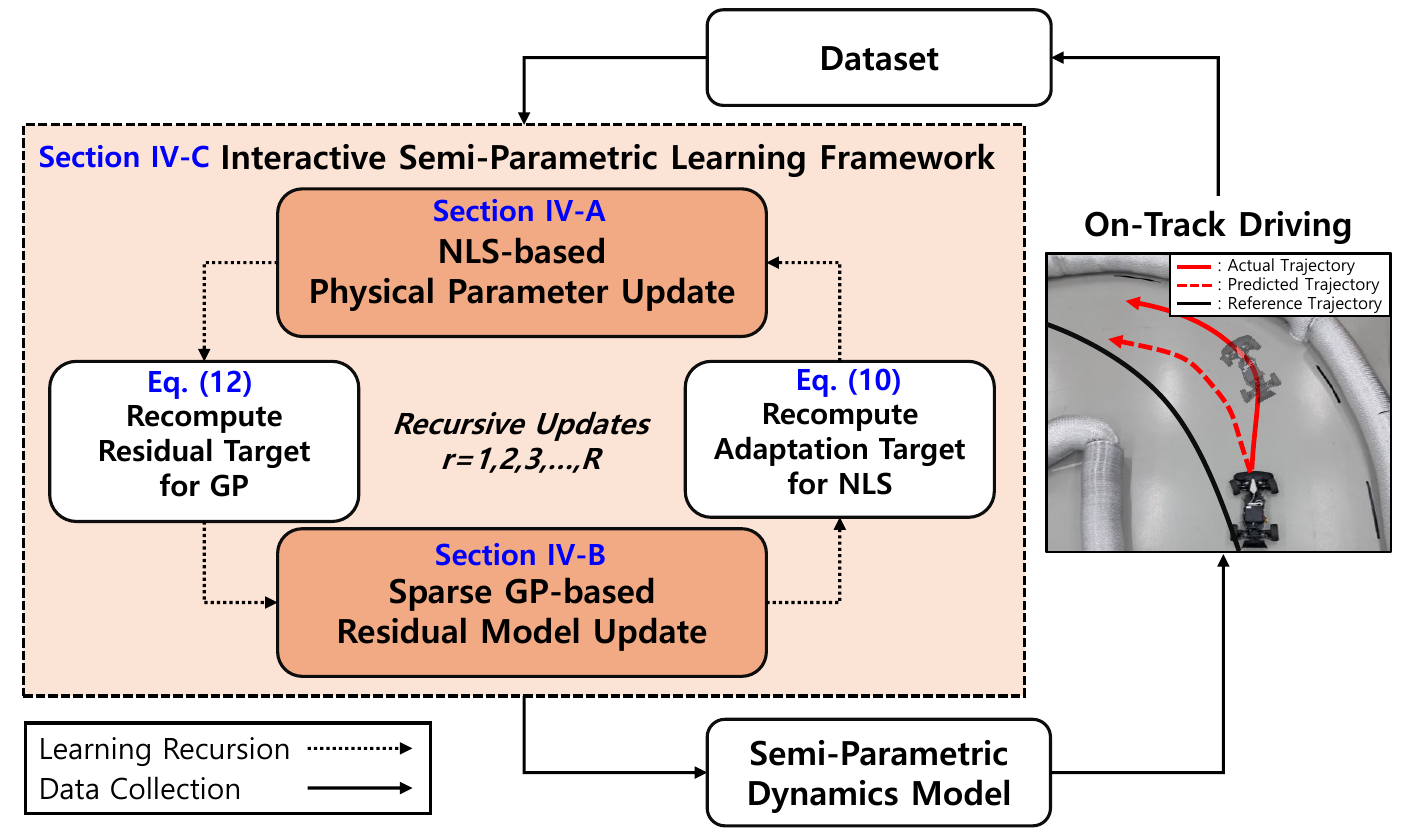}
    \caption{Overall architecture of the proposed interactive learning framework.}
    \label{fig:algorithm}
\end{figure}

Given a dataset $\mathcal{D} := \{(\mathbf{x}_k,\mathbf{u}_k, \mathbf{x}_{k+1}^{\mathrm{meas}})\}_{k=0}^{N-1}$, where $\mathbf{x}_{k+1}^{\mathrm{meas}}$ denotes the measured next state, we define $J_{\mathcal{D}}$ as the weighted mean squared error (MSE) of the one-step prediction of~\eqref{eq.semiparametric}:
\begin{equation}\label{eq.joint_loss}
J_{\mathcal{D}}(\theta,g) :=
\frac{1}{N}\sum_{k=0}^{N-1} \left\| WS\left(
\mathbf{x}_{k+1}^{\mathrm{meas}} - f(\mathbf{x}_k,\mathbf{u}_k;\theta) - g(\mathbf{z}_k)
\right) \right\|_2^2,
\end{equation}
where $S:\mathbb{R}^{6}\to\mathbb{R}^{2}$ is a linear map that extracts $(v_y,\omega)$ from a state vector, consistent with the residual model, and $W\in\mathbb{R}^{2\times 2}$ is a weighting matrix.
However, directly solving~\eqref{eq.interactive_problem} is challenging because it involves two variables of different nature: a finite-dimensional parameter vector $\theta$ that enters $f$ nonlinearly and a nonparametric function $g$.
We therefore adopt a recursive optimization strategy that alternately updates $\theta$ and $g$, each while holding the other fixed:
\begin{align}
\theta^{(r)} &= \arg\min_{\theta \in \Theta}\; J_{\mathcal{D}}\big(\theta, g^{(r-1)}\big), \label{eq.theta_update}\\
g^{(r)} &= \arg\min_{g \in \mathcal{G}}\; J_{\mathcal{D}}\big(\theta^{(r)}, g\big), \label{eq.g_update}
\end{align}
where $r=1,\dots,R$ denotes the recursion step, and the recursion starts from the nominal parameters $\theta^{(0)}=\theta_0$ and $g^{(0)}\equiv\mathbf{0}$.
The parameter update~\eqref{eq.theta_update} is a standard NLS problem, whereas the residual model update~\eqref{eq.g_update} is handled by GP regression.
In this way, the parametric and residual models are learned interactively to improve prediction accuracy.
The overall architecture of the proposed framework is illustrated in Fig.~\ref{fig:algorithm}, with the algorithmic details and theoretical analysis presented in the following section.

\section{Algorithm Development}

\subsection{NLS-based Physical Parameter Update}\label{sec:param}
The parameter update in~\eqref{eq.theta_update} aims to estimate the physical parameters while accounting for the current residual model.
At recursion step $r$, we define the adaptation target as
\begin{equation}\label{eq.nls_target}
\tilde{\mathbf{x}}^{(r)}_{k+1} := \mathbf{x}^{\mathrm{meas}}_{k+1} - g^{(r-1)}(\mathbf{z}_k).
\end{equation}
Substituting~\eqref{eq.nls_target} into~\eqref{eq.theta_update} yields the following weighted NLS problem:
\begin{equation}\label{eq.param_nls}
\theta^{(r)} = \arg\min_{\theta\in\Theta}
\frac{1}{N}\sum_{k=0}^{N-1}
\left\| WS\left( \tilde{\mathbf{x}}_{k+1}^{(r)} - f(\mathbf{x}_k,\mathbf{u}_k;\theta) \right) \right\|_2^2.
\end{equation}
The resulting nonconvex optimization problem is solved over $\Theta$ using a numerical solver warm-started from $\theta^{(r-1)}$. If the solver returns a larger objective value than that of $\theta^{(r-1)}$, we retain $\theta^{(r-1)}$.
The updated parameters $\theta^{(r)}$ are subsequently used to recompute the residual targets for GP learning.

\subsection{Sparse GP-based Residual Model Update}\label{sec:gp}
With the updated parameters $\theta^{(r)}$ fixed, the residual model is then learned to capture the mismatch left unexplained by the parametric model.
Replacing the actual next states in~\eqref{eq.residual_target} with the measured ones, we compute the residual target as
\begin{equation}\label{eq.residual_def}
\mathbf{y}_{g,k}^{(r)} := S\left(
\mathbf{x}_{k+1}^{\mathrm{meas}} -
f(\mathbf{x}_k,\mathbf{u}_k;\theta^{(r)}) \right).
\end{equation}
Based on these targets, we train $g_{v_y}$ and $g_\omega$ as two independent GPs over $\mathbf{z}$, with radial basis function (RBF) kernels $k_{v_y}$ and $k_\omega$ and Gaussian observation noise.

To reduce computational complexity, we employ a sparse GP approximation with $M$ inducing points.
Two common approaches to training sparse GPs are the fully independent training conditional (FITC) approximation~\cite{fitc} and the variational approach based on the evidence lower bound (ELBO)~\cite{sparsegp}.  
We adopt the variational approach, which has been shown to yield more accurate predictions than FITC~\cite{sparsegp_mpc}.
Accordingly, each GP is trained by minimizing the negative ELBO via gradient-based optimization~\cite{svgp}, warm-started from the previous model $g^{(r-1)}$ at each recursion step.
The posterior means of the two GPs then form the residual model $g$ in~\eqref{eq.semiparametric}.

Standard sparse GP training, however, does not explicitly account for two aspects of our interactive framework. 
First, the residual targets change with each parameter update. 
Second, the GP training objective differs from the shared prediction-error objective $J_{\mathcal{D}}$. 
To address these two aspects, we introduce an adaptive learning rate and min-MSE model selection, respectively.

\textbf{1) Adaptive Learning Rate:}
Since GP training is warm-started from $g^{(r-1)}$, the size of the required GP update depends on how much the residual target has changed.
A large change may require faster GP learning, whereas a small change allows finer refinement.
Motivated by this, we adjust the GP learning rate using the magnitude of the parameter update as a proxy for the change in the residual target:
\begin{equation}\label{eq.gp_lr}
\eta^{(r)} = \eta_0 + (\eta_1-\eta_0)
\frac{\Delta\theta^{(r)}}{\Delta\theta_{\max}^{(r)}}, \qquad r\geq 2,
\end{equation}
where
\begin{equation*}
\Delta\theta^{(r)}
:= \|\theta^{(r)}-\theta^{(r-1)}\|,
\qquad
\Delta\theta_{\max}^{(r)}
:= \max_{2\leq j\leq r}\Delta\theta^{(j)}.
\end{equation*}
Here, $\eta_0$ and $\eta_1$ denote the minimum and maximum learning rates, respectively, and $\eta^{(1)}=\eta_1$ since no previous residual target exists at $r=1$.\footnote{If $\Delta\theta_{\max}^{(r)}=0$, the parameters have not changed and neither has the residual target, so we set $\eta^{(r)}=\eta_0$.}
The running maximum normalizes the parameter changes and ensures that $\eta^{(r)}\in[\eta_0,\eta_1]$, allowing the learning rate to decrease as subsequent parameter updates become smaller.

\textbf{2) Min-MSE Model Selection:}
To align the GP update with the shared objective~\eqref{eq.joint_loss}, we select the candidate with the lowest MSE among those obtained during GP training:
\begin{equation}\label{eq.gp_selection}
g^{(r)}
=
\arg\min_{g\in\mathcal{G}_{\mathrm{cand}}^{(r)}}
\frac{1}{N}\sum_{k=0}^{N-1}
\left\|
W\left(\mathbf{y}_{g,k}^{(r)}
-
S g(\mathbf{z}_k)\right)
\right\|_2^2,
\end{equation}
where $\mathcal{G}_{\mathrm{cand}}^{(r)}\subseteq\mathcal{G}$ denotes the set of these candidate models, including the previous model $g^{(r-1)}$.

\subsection{Interactive On-Track Learning Framework}\label{sec:framework}
The recursion in~\eqref{eq.theta_update} and~\eqref{eq.g_update} is embedded in a learning procedure that interactively updates the parametric and residual models from on-track data:
\begin{enumerate}
\item \textit{Initial data collection:} The nominal model $f(\cdot;\theta_0)$ is deployed on the track to collect an initial dataset $\mathcal{D}$.

\item \textit{Recursive update:} Given $\mathcal{D}$, the recursion is initialized with $\theta^{(0)}=\theta_0$ and $g^{(0)}\equiv\mathbf{0}$, and each recursion step $r=1,\dots,R$ updates the two models in turn, each using the latest update of the other.
First, the parameters are updated to $\theta^{(r)}$ by solving~\eqref{eq.param_nls} with the adaptation target~\eqref{eq.nls_target}.
Second, the residual model is updated to $g^{(r)}$ by training the sparse GP on the residual target~\eqref{eq.residual_def} with the learning rate~\eqref{eq.gp_lr} and selecting the model by~\eqref{eq.gp_selection}.

\item \textit{Deployment:} After $R$ recursion steps, the final pair $(\theta^{(R)},g^{(R)})$ is deployed, and the resulting semi-parametric model~\eqref{eq.semiparametric} is used in a model-based tracking controller on the track.
\end{enumerate}
Steps~2 and~3 can be repeated by aggregating the data collected during deployment into $\mathcal{D}$.

\subsection{Theoretical Analysis}\label{sec:theoretical_analysis}

\begin{figure}[!t]
    \centering
    \includegraphics[width=0.45\textwidth]{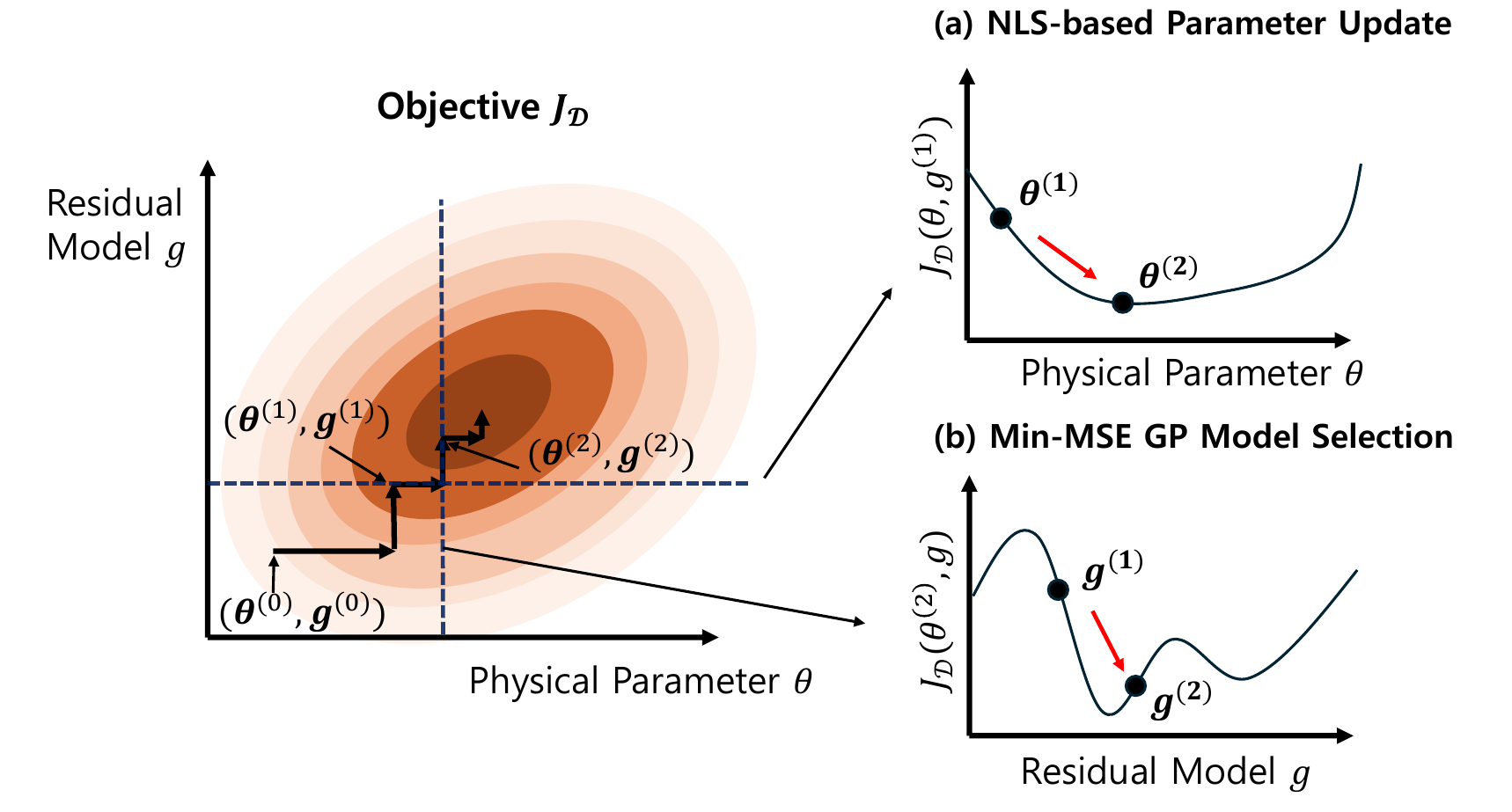}
    \caption{Schematic illustration of the recursion as approximate block-coordinate optimization of $J_{\mathcal{D}}$: (a) NLS-based parameter update with fallback to $\theta^{(r-1)}$ and (b) min-MSE model selection among GP candidates both guarantee that $J_{\mathcal{D}}$ does not increase.}
    \label{fig:bcd}
\end{figure}

The recursion in~\eqref{eq.theta_update} and~\eqref{eq.g_update} can be interpreted as an approximate block-coordinate optimization~\cite{bcd} of the shared prediction-error objective~\eqref{eq.joint_loss}, as illustrated in Fig.~\ref{fig:bcd}.
The NLS solver approximately solves the nonconvex parameter-update problem~\eqref{eq.theta_update}, while the GP update minimizes $J_{\mathcal{D}}(\theta^{(r)},\cdot)$ over the candidate set $\mathcal{G}_{\mathrm{cand}}^{(r)}$.
Nevertheless, a monotonically non-increasing objective requires only that neither update increases $J_{\mathcal{D}}$, as formalized below.

\begin{theorem}[Monotonic convergence of the prediction-error objective]\label{thm:monotone}
Consider the recursion in Section~\ref{sec:framework} on a fixed dataset $\mathcal{D}$, generating $\{(\theta^{(r)},g^{(r)})\}_{r\geq 0}$.
Suppose that (i) the parameter update retains $\theta^{(r-1)}$ whenever the solver returns a larger objective value, and (ii) $g^{(r-1)}\in\mathcal{G}_{\mathrm{cand}}^{(r)}$ for all $r\geq 1$.
Then the sequence $\{J_{\mathcal{D}}(\theta^{(r)},g^{(r)})\}_{r\geq 0}$ is monotonically non-increasing and converges to a finite limit.
\end{theorem}

\begin{proof}
By (i), $\theta^{(r)}$ is either a solver output whose objective does not exceed that of $\theta^{(r-1)}$, or $\theta^{(r-1)}$ itself, so that
\begin{equation}\label{eq.theta_descent}
J_{\mathcal{D}}(\theta^{(r)},g^{(r-1)}) \leq J_{\mathcal{D}}(\theta^{(r-1)},g^{(r-1)}).
\end{equation}
By (ii), and since the criterion in~\eqref{eq.gp_selection} equals $J_{\mathcal{D}}(\theta^{(r)},g)$, the selected model satisfies
\begin{equation}\label{eq.g_descent}
J_{\mathcal{D}}(\theta^{(r)},g^{(r)}) \leq J_{\mathcal{D}}(\theta^{(r)},g^{(r-1)}).
\end{equation}
Chaining~\eqref{eq.theta_descent} and~\eqref{eq.g_descent} gives $J_{\mathcal{D}}(\theta^{(r)},g^{(r)})\leq J_{\mathcal{D}}(\theta^{(r-1)},g^{(r-1)})$ for all $r\geq 1$.
The sequence is therefore monotonically non-increasing, and since $J_{\mathcal{D}}\geq 0$, it converges to a finite limit.
\end{proof}
Consequently, the final pair $(\theta^{(R)},g^{(R)})$ attains the lowest objective among all recursion steps, which justifies deploying it without a separate selection step.

\section{Numerical Simulation}
\subsection{Simulation Setup} \label{sec:sim_setup}
The ground-truth vehicle dynamics are simulated using a double-track model with horizontal and vertical shifts added to the simplified Pacejka law~\cite{param_adap_local_min}.
In each of 20 scenarios on the ETH Zurich race track~\cite{param_adap_lla}, the tire parameters and yaw inertia of the ground-truth model are uniformly sampled from $B,C\in[1.0,1.4]$, $\mu\in[0.4,0.8]$, and $I_z\in[0.014,0.024]\,\mathrm{kg\!\cdot\!m^2}$.
The proposed framework instead builds on the single-track parametric model~\eqref{eq.parametric} with $m=3.0\,\mathrm{kg}$ and $L_f=L_r=0.14\,\mathrm{m}$, whose nominal parameters $\theta_0$ are set to the midpoints of these ranges, i.e., $B=1.2$, $C=1.2$, $\mu=0.6$, and $I_z=0.019\,\mathrm{kg\!\cdot\!m^2}$.
This setup therefore introduces both parameter and model-structure mismatches, allowing us to evaluate the effects of parametric and nonparametric learning.

The reference trajectory is obtained by solving a minimum-lap-time optimization problem~\cite{Nominal} using the ground-truth dynamics rather than the nominal model.
This choice removes model mismatch from reference generation, so that lap-time differences reflect tracking performance more directly.
A tracking model predictive controller (MPC) is implemented using \texttt{FORCESPRO}~\cite{forcespro} with a sampling time of $0.05\,\mathrm{s}$. 
The dataset $\mathcal{D}$ is collected from the second lap of a single run with the nominal model.
The NLS problem~\eqref{eq.param_nls} is solved using \texttt{Ipopt}~\cite{ipopt} over $\Theta$, which is set to the sampling ranges above.
The residual model $g$ is learned using a sparse GP~\cite{svgp} with $M=10$. 
The weighting matrix in~\eqref{eq.joint_loss} is set to $W=\mathrm{diag}(1,1)$, and the number of recursion steps to $R=6$.\footnote{A GitHub link with detailed parameters will be provided in the final version.} 

\begin{figure}[!t]
    \centering
    \includegraphics[width=\columnwidth]{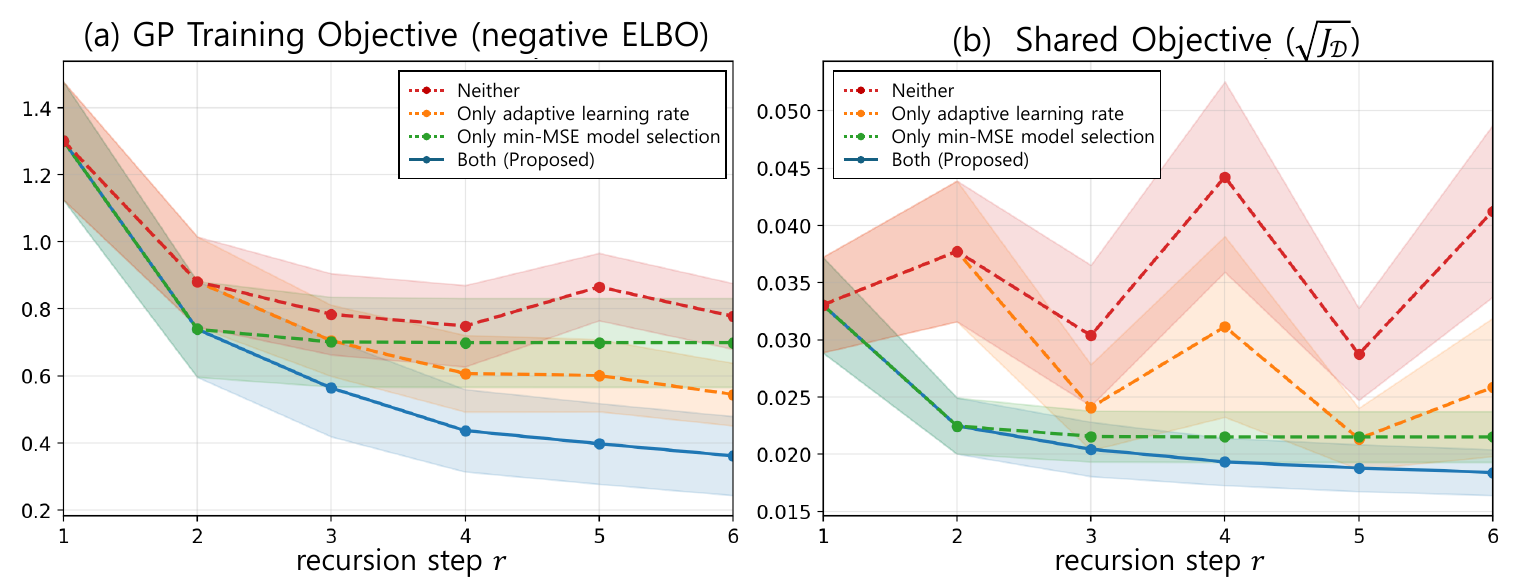}
    \caption{Ablation of the sparse GP learning strategy over recursion steps $r$, with $\pm0.5$ standard-deviation shading: (a) Negative ELBO and (b) $\sqrt{J_{\mathcal{D}}}$.}
    \label{fig:ablation}
\end{figure}

\begin{table}[t]
\centering
\caption{Ablation of the sparse GP learning strategy over the 20 simulation scenarios. Values are mean $\pm$ one standard deviation at the final recursion step.}
\label{tab:ablation}
\resizebox{\linewidth}{!}{%
\begin{tabular}{lcc}
\hline
 & Final Negative ELBO & Final $\sqrt{J_{\mathcal{D}}}$\\
\hline
Neither & $0.7780 \pm 0.1961$ & $0.0412 \pm 0.0150$ \\
Only adaptive learning rate & $0.5448 \pm 0.1865$ & $0.0259 \pm 0.0121$ \\
Only min-MSE model selection & $0.6990 \pm 0.2645$ & $0.0215 \pm 0.0044$ \\
\textbf{Both (Proposed)} & $\mathbf{0.3616 \pm 0.2357}$ & $\mathbf{0.0184 \pm 0.0040}$ \\
\hline
\end{tabular}
}
\end{table}

\subsection{Ablation of the Sparse GP Learning Strategy}
We first examine the two designs of the sparse GP learning strategy introduced in Section~\ref{sec:gp}: the adaptive learning rate and min-MSE model selection.
To isolate their effects, we compare four configurations of the proposed framework, which use neither design, only the adaptive learning rate, only min-MSE model selection, or both designs.
For each configuration, we evaluate two metrics: (i) the negative ELBO, which represents the GP training objective; and (ii) $\sqrt{J_{\mathcal{D}}}$, which expresses the shared objective~\eqref{eq.joint_loss} as a root-mean-square error (RMSE).
Fig.~\ref{fig:ablation} shows both metrics over $R=6$ recursion steps, and Table~\ref{tab:ablation} reports their values at the final step.

With the adaptive learning rate only, the negative ELBO continues to decrease, whereas $\sqrt{J_{\mathcal{D}}}$ fluctuates across recursion steps, showing that a lower negative ELBO does not imply a lower $J_{\mathcal{D}}$.
With min-MSE model selection only, $\sqrt{J_{\mathcal{D}}}$ decreases monotonically but levels off after the second recursion step.
With both designs, $\sqrt{J_{\mathcal{D}}}$ keeps decreasing monotonically beyond the second recursion step, and both metrics reach their lowest final mean values.
These results suggest that the adaptive learning rate facilitates continued GP refinement, while min-MSE model selection prevents increases in $J_{\mathcal{D}}$ during the GP update, consistent with Theorem~\ref{thm:monotone}.

\begin{figure*}[!t]
    \centering
    \includegraphics[width=\textwidth]{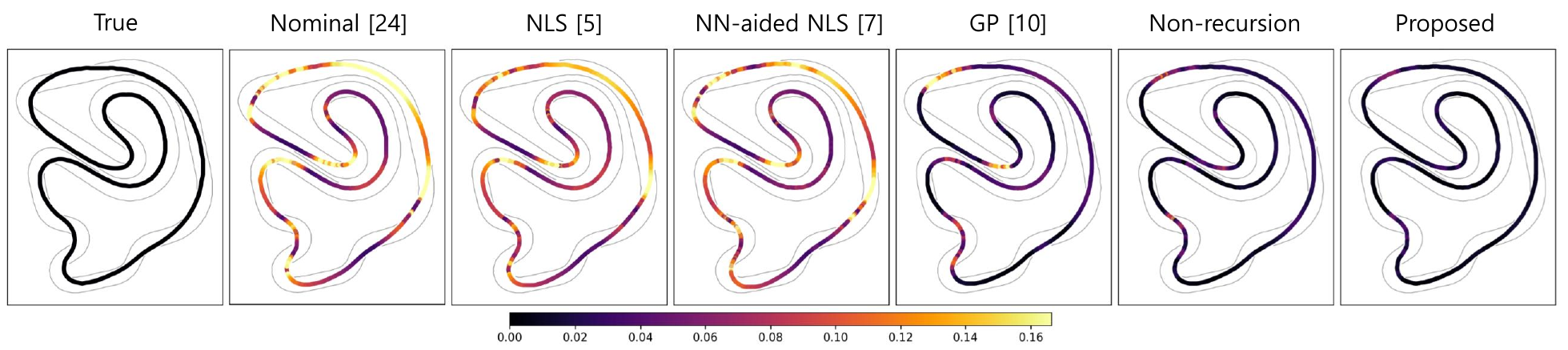}
    \caption{Sum of one-step prediction errors in $v_y$ and $\omega$ mapped along the actual trajectories in a representative simulation scenario.}
    \label{fig:error_map}
\end{figure*}

\begin{table*}[!t]
\centering
\caption{Performance comparison across 20 numerical simulation scenarios (mean $\pm$ one standard deviation).}
\label{tab:comparison}
\resizebox{0.9\textwidth}{!}{%
\begin{tabular}{|l|c|c|c|c|c|}
\hline
\multirow{2}{*}{Method}
& \multicolumn{2}{c|}{One-step prediction error}
& \multicolumn{2}{c|}{Tracking error}
& Lap time \\ \cline{2-6}

& $\mathcal{E}_{v_y}$ [m/s]
& $\mathcal{E}_{\omega}$ [rad/s]
& $\mathrm{RMSE}_{e_y}$ [m]
& $\mathrm{RMSE}_{v_x}$ [m/s]
& $T$ [s] \\ \hline

True
& $0.0000 \pm 0.0000$
& $0.0000 \pm 0.0000$
& $0.0609 \pm 0.0265$
& $0.0334 \pm 0.0130$
& $29.55 \pm 1.04$ \\ \hline

Nominal~\cite{Nominal}
& $0.0309 \pm 0.0083$
& $0.0899 \pm 0.0134$
& $0.2478 \pm 0.0214$
& $0.1684 \pm 0.0398$
& $31.03 \pm 0.95$ \\ \hline

NLS~\cite{param_adap_NLS}
& $0.0267 \pm 0.0038$
& $0.0765 \pm 0.0108$
& $0.2231 \pm 0.0092$
& $0.1790 \pm 0.0204$
& $30.86 \pm 1.06$ \\ \hline

NN-aided NLS~\cite{param_adap_NN}
& $0.0304 \pm 0.0066$
& $0.1081 \pm 0.0397$
& $0.2465 \pm 0.0220$
& $0.2117 \pm 0.0318$
& $31.10 \pm 0.86$ \\ \hline

GP~\cite{residual_gp}
& $0.0086 \pm 0.0020$
& $0.0506 \pm 0.0231$
& $0.1834 \pm 0.0799$
& $0.1028 \pm 0.0504$
& $30.51 \pm 1.42$ \\ \hline

Non-recursion
& $0.0039 \pm 0.0009$
& $0.0308 \pm 0.0095$
& $0.1553 \pm 0.0653$
& $0.0917 \pm 0.0480$
& $30.28 \pm 1.35$ \\ \hline

\textbf{Proposed}
& $\mathbf{0.0031 \pm 0.0005}$
& $\mathbf{0.0165 \pm 0.0025}$
& $\mathbf{0.1195 \pm 0.0571}$
& $\mathbf{0.0645 \pm 0.0417}$
& $\mathbf{29.99 \pm 1.25}$ \\ \hline

\end{tabular}
}
\end{table*}

\subsection{Comparative Analysis}
To evaluate the effectiveness of the proposed framework, the following baselines are implemented within the same tracking MPC with different dynamics models:
\begin{itemize}
    \item \textbf{Nominal}~\cite{Nominal}: Uses the parametric model with $\theta_0$, without learning or a residual model.
    \item \textbf{NLS}~\cite{param_adap_NLS}: Adapts only $\theta$ through NLS, without a residual model. 
    \item \textbf{NN-aided NLS}~\cite{param_adap_NN}: Adapts only $\theta$ through NLS, using additional training data simulated with an NN-based residual model. The NN is used only to generate these data, not in the MPC.
    \item \textbf{GP}~\cite{residual_gp}: Learns a GP-based residual model, with the physical parameters fixed at $\theta_0$.
\end{itemize}
To examine the effect of recursion, we additionally evaluate the following variant of the proposed framework:
\begin{itemize}
    \item \textbf{Non-recursion}: Performs a single NLS update followed by a single GP update ($R=1$).
\end{itemize}
In each scenario, all learned models are trained on the same dataset $\mathcal{D}$. For reference, we also report \textbf{True}, in which the MPC uses the ground-truth dynamics.
For each method, dynamics prediction accuracy is evaluated using the one-step prediction RMSEs in lateral velocity and yaw rate, $\mathcal{E}_{v_y}$ and $\mathcal{E}_{\omega}$, respectively.
These errors are computed between the next states predicted by the MPC and those measured during deployment.
Tracking performance is assessed using the RMSEs of lateral-position and speed deviations from the reference trajectory, $\mathrm{RMSE}_{e_y}$ and $\mathrm{RMSE}_{v_x}$, respectively, along with the lap time $T$.
Fig.~\ref{fig:error_map} shows a representative scenario, and Table~\ref{tab:comparison} summarizes the results over the 20 scenarios.

The results show that \emph{Nominal}~\cite{Nominal} yields large prediction errors in both $v_y$ and $\omega$.
\emph{NLS}~\cite{param_adap_NLS} reduces both errors, but adapting $\theta$ alone cannot compensate for the unmodeled effects of the double-track dynamics.
\emph{NN-aided NLS}~\cite{param_adap_NN} performs worse than \emph{NLS}, even with its augmented training data.
Since the simulated data are generated under the model-structure mismatch, they deviate from the ground-truth dynamics and bias the adapted parameters.
\emph{GP}~\cite{residual_gp} reduces the prediction errors further, but its $\mathcal{E}_{\omega}$ remains high, likely because its bounded inducing-point capacity must also compensate for the parameter mismatch.
In contrast, \emph{Non-recursion} and \emph{Proposed} outperform all baselines in every metric, showing the benefit of combining parametric and nonparametric learning.
\emph{Non-recursion}, nonetheless, leaves noticeable errors in several track sections, as shown in Fig.~\ref{fig:error_map}.
Through the recursion, \emph{Proposed} further reduces $\mathcal{E}_{\omega}$ by 46.4\% relative to \emph{Non-recursion}.
Overall, it reduces $\mathcal{E}_{v_y}$ and $\mathcal{E}_{\omega}$ by 90.0\% and 81.6\% relative to \emph{Nominal}. 
As a result, it attains the smallest tracking errors and the shortest lap time among the learned models, approaching those of \emph{True}.
These results highlight the benefit of interactive learning given the same limited on-track data from a single lap.

\section{Hardware Experiment}
\begin{figure}[!t]
    \centering
    \includegraphics[width=\columnwidth]{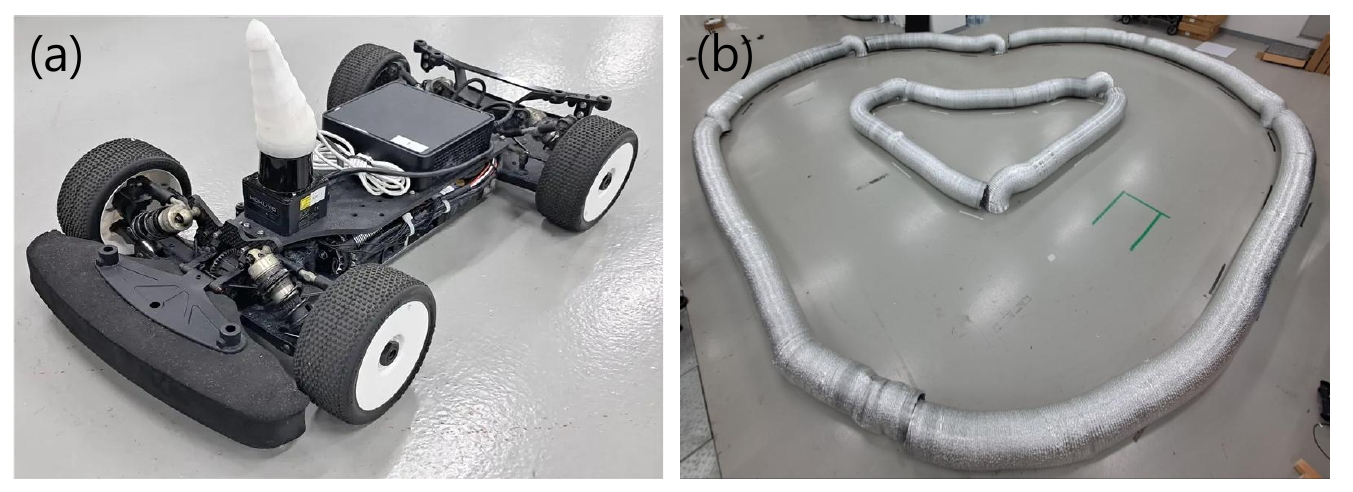}
    \caption{Experimental setup: (a) 1/10-scale vehicle and (b) indoor track.}
    \label{fig:experiment_setup}
\end{figure}

The proposed framework is further evaluated on a 1/10-scale autonomous racing platform on an indoor track, shown in Fig.~\ref{fig:experiment_setup}.
The vehicle is equipped with an onboard Intel NUC computer, and its states are estimated by fusing LiDAR, IMU, and wheel odometry.
Since the actual dynamics are unknown in hardware experiments, the reference trajectory is optimized using the nominal model and kept identical across all methods.
All other settings and the evaluation metrics are the same as in simulation, with the one-step prediction errors computed with respect to the estimated states.
Each learned model is deployed in five separate runs, and its performance is evaluated on the second lap of each run to exclude the initial acceleration phase.

\begin{figure*}[!t]
    \centering
    \includegraphics[width=0.92\textwidth]{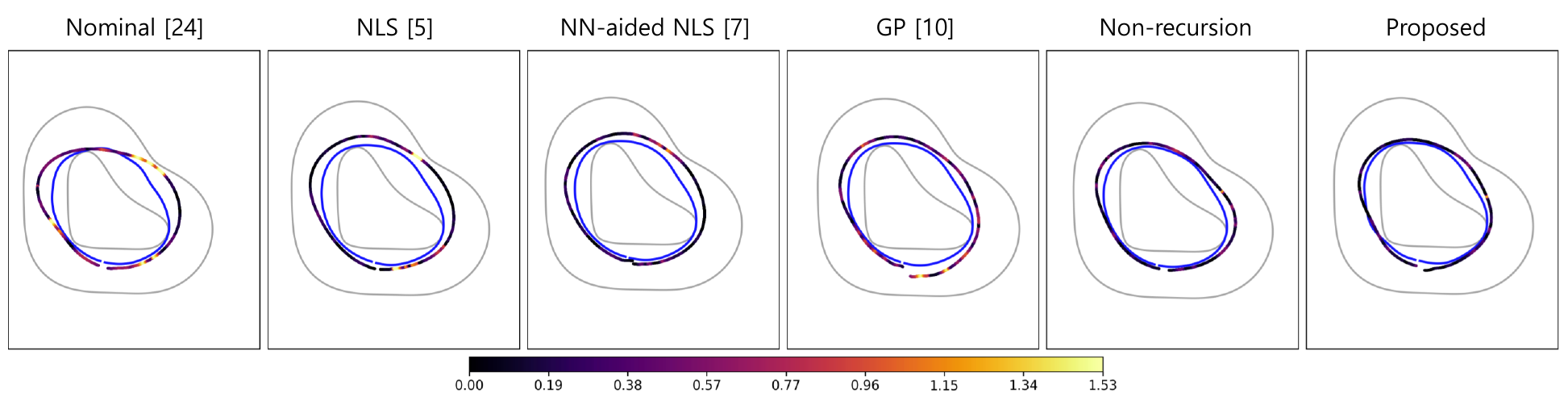}
    \caption{Sum of one-step prediction errors in $v_y$ and $\omega$ mapped along the actual trajectory in a representative hardware run. The blue line denotes the reference trajectory.}
    \label{fig:hw_error_map}
\end{figure*}

Table~\ref{tab:hardware} summarizes the hardware results, and Fig.~\ref{fig:hw_error_map} shows the one-step prediction errors mapped along the actual trajectory in a representative run.
As in simulation, \emph{Non-recursion} and \emph{Proposed} outperform all baselines in every metric, showing the benefit of learning both models on the real vehicle.
Through the recursion, \emph{Proposed} further reduces $\mathcal{E}_{\omega}$ by 18.1\% relative to \emph{Non-recursion} and shortens the lap time by $0.17\,\mathrm{s}$.
Overall, it reduces $\mathcal{E}_{v_y}$ and $\mathcal{E}_{\omega}$ by 54.0\% and 59.4\% and the lap time by $0.48\,\mathrm{s}$ relative to \emph{Nominal}.
Consistent with this, it achieves the lowest lateral tracking error, reducing $\mathrm{RMSE}_{e_y}$ by 42.4\% relative to \emph{Nominal}, with closer tracking of the reference trajectory as illustrated in Fig.~\ref{fig:hw_error_map}.
These results indicate that the advantages observed in simulation persist under real-world uncertainty.

\begin{table*}[!t]
\centering
\caption{Performance comparison in hardware experiments (mean $\pm$ one standard deviation over five runs).}
\label{tab:hardware}
\resizebox{0.9\textwidth}{!}{%
\begin{tabular}{|l|c|c|c|c|c|}
\hline
\multirow{2}{*}{Method} & \multicolumn{2}{c|}{One-step prediction error} & \multicolumn{2}{c|}{Tracking error} & Lap time \\ \cline{2-6}
 & $\mathcal{E}_{v_y}$ [m/s] & $\mathcal{E}_{\omega}$ [rad/s] & $\mathrm{RMSE}_{e_y}$ [m] & $\mathrm{RMSE}_{v_x}$ [m/s] & $T$ [s] \\ \hline
Nominal~\cite{Nominal}& $0.0615 \pm 0.0035$ & $0.3844 \pm 0.0171$ & $0.2976 \pm 0.0035$ & $0.3862 \pm 0.0109$ & $5.890 \pm 0.119$ \\ \hline
NLS~\cite{param_adap_NLS} & $0.0438 \pm 0.0038$ & $0.2278 \pm 0.0284$ & $0.2909 \pm 0.0034$ & $0.3379 \pm 0.0094$ & $5.800 \pm 0.035$ \\ \hline
NN-aided NLS~\cite{param_adap_NN} & $0.0409 \pm 0.0045$ & $0.2087 \pm 0.0308$ & $0.2725 \pm 0.0035$ & $0.3374 \pm 0.0045$ & $5.740 \pm 0.022$ \\ \hline
GP~\cite{residual_gp} & $0.0385 \pm 0.0032$ & $0.3417 \pm 0.0356$ & $0.2333 \pm 0.0076$ & $0.3305 \pm 0.0051$ & $5.676 \pm 0.048$ \\ \hline
Non-recursion & $0.0323 \pm 0.0031$ & $0.1903 \pm 0.0437$ & $0.2066 \pm 0.0126$ & $0.3229 \pm 0.0139$ & $5.578 \pm 0.056$ \\ \hline
\textbf{Proposed} & $\mathbf{0.0283 \pm 0.0040}$ & $\mathbf{0.1559 \pm 0.0398}$ & $\mathbf{0.1713 \pm 0.0138}$ & $\mathbf{0.3199 \pm 0.0111}$ & $\mathbf{5.410 \pm 0.060}$ \\ \hline
\end{tabular}
}
\end{table*}

\section{Conclusion}
This paper has presented an interactive framework for semi-parametric dynamics learning from on-track data in autonomous racing.
The framework recursively updates the physical parameters via NLS and the residual model via sparse GP regression in turn under a single shared prediction-error objective.
To account for their mutual dependence, each update is guided by the latest update of the other model.
Since an update of either model is accepted only if it does not increase the shared objective, the objective never increases and is guaranteed to converge over the recursion.
Numerical simulations and hardware experiments on a 1/10-scale vehicle showed that the proposed framework outperforms the baselines, given the same limited on-track data.
Future work will extend the framework to online learning from streaming on-track data, to account for tire degradation and varying track surface conditions during racing.
\bibliographystyle{IEEEtran}
\bibliography{reference}

\end{document}